\documentclass{article}

\usepackage{arxiv}

\usepackage[utf8]{inputenc} 
\usepackage[T1]{fontenc}    
\usepackage{url}            
\usepackage{booktabs}       
\usepackage{amsfonts}       
\usepackage{nicefrac}       
\usepackage{microtype}      
\usepackage{lipsum}
\usepackage{fancyhdr}       
\usepackage{graphicx}       
\graphicspath{{media/}}     
\usepackage{amsmath}
\usepackage{bbding}
\usepackage[ruled]{algorithm2e}
\usepackage{soul}

\usepackage{colortbl}
\usepackage[most]{tcolorbox}
\definecolor{citeblue}{HTML}{5B9BD5}
\definecolor{bright_purple}{HTML}{7B1B6D}
\definecolor{dark_pink}{HTML}{A93652}
\usepackage[colorlinks=true]{hyperref} 

\hypersetup{
    citecolor=citeblue,   
    linkcolor=red,   
    urlcolor=citeblue     
}

\title{SeqTex: Generate Mesh Textures in Video Sequence} 

\author{
  Ze Yuan,$^{1\star}$ 
  Xin Yu,$^{1\star\dagger}$ 
  Yangtian Sun,$^{1}$ 
  Yuan-Chen Guo,$^{2}$ 
  Yan-Pei Cao,$^{2}$ 
  Ding Liang,$^{2}$ 
  Xiaojuan Qi$^{1\ddagger}$ \\ 
  \\
  $^{1}$HKU, $^{2}$VAST \\
  \textsuperscript{$\star$}Equal contribution, 
  \textsuperscript{$\dagger$}Project lead,
  \textsuperscript{$\ddagger$}Corresponding author
}

\begin{document}
\maketitle

\begin{figure}[htbp]
    \centering
    \vspace{-4em}
    \includegraphics[width=\textwidth]{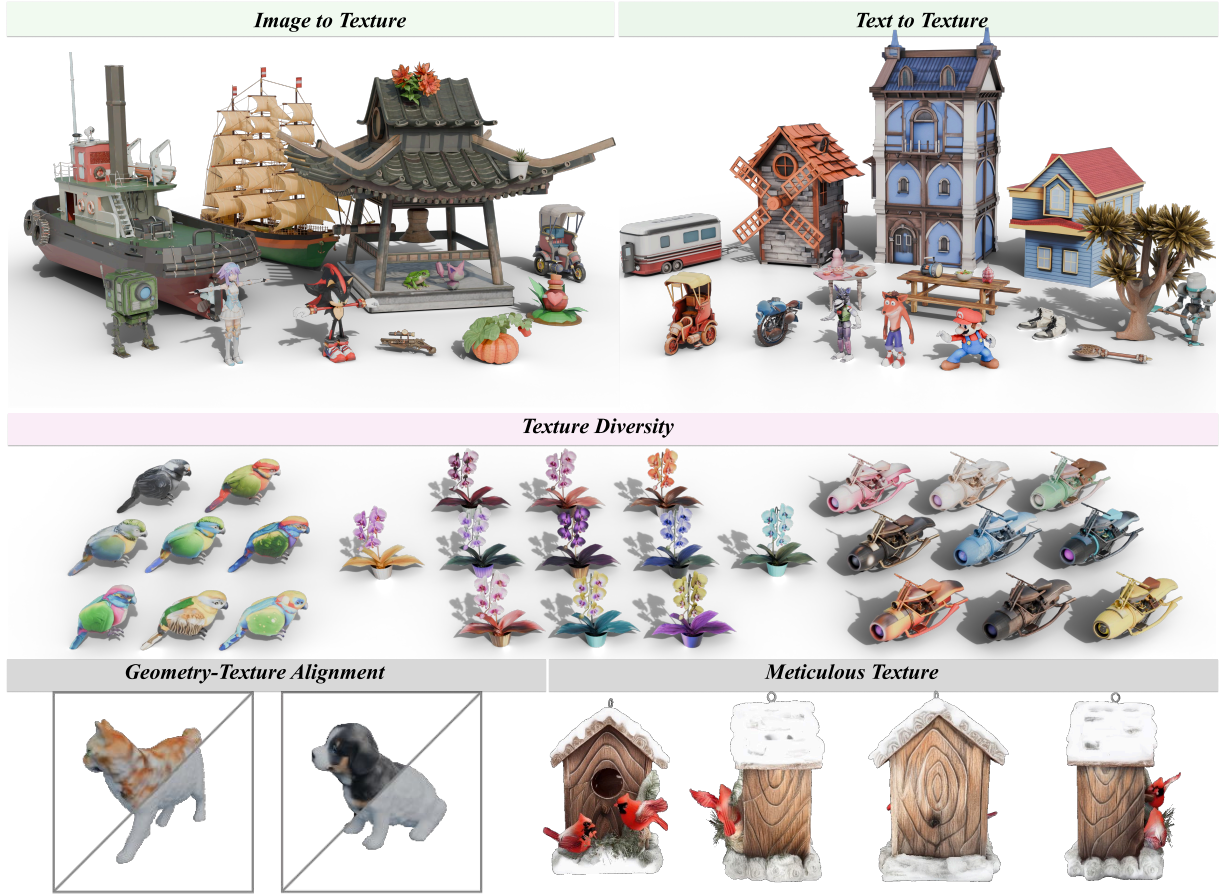}
    \caption{ 
    {SeqTex is capable of generating high-quality and diverse textures for meshes, conditioned on either an image or text input. Furthermore, the generated textures maintain a strong correspondence with the mesh's geometry and ensure high consistency of detailed textures across various viewing angles.}
    }
\label{fig:teaser}
\end{figure}

\begin{abstract}
Training native 3D texture generative models remains a fundamental yet challenging problem, largely due to the limited availability of large-scale, high-quality 3D texture datasets. This scarcity hinders generalization to real-world scenarios. To address this, most existing methods finetune foundation image generative models to exploit their learned visual priors. However, these approaches typically generate only multi-view images and rely on post-processing to produce UV texture maps—an essential representation in modern graphics pipelines. Such two-stage pipelines often suffer from error accumulation and spatial inconsistencies across the 3D surface.  
In this paper, we introduce \textbf{SeqTex}, a novel end-to-end framework that leverages the visual knowledge encoded in pretrained video foundation models to directly generate complete UV texture maps. Unlike previous methods that model the distribution of UV textures in isolation, \textbf{SeqTex} reformulates the task as a sequence generation problem, enabling the model to learn the joint distribution of multi-view renderings and UV textures. This design effectively transfers the consistent image-space priors from video foundation models into the UV domain.
To further enhance performance, we propose several architectural innovations: a decoupled multi-view and UV branch design, geometry-informed attention to guide cross-domain feature alignment, and adaptive token resolution to preserve fine texture details while maintaining computational efficiency. Together, these components allow \textbf{SeqTex} to fully utilize pretrained video priors and synthesize high-fidelity UV texture maps without the need for post-processing.
Extensive experiments show that SeqTex achieves state-of-the-art performance on both image-conditioned and text-conditioned 3D texture generation tasks, with superior 3D consistency, texture-geometry alignment, and real-world generalization. Our project page is \href{https://yuanze1024.github.io/SeqTex/}{https://yuanze1024.github.io/SeqTex/}.

\end{abstract}



\section{Introduction}
Manually texturing 3D models is a highly labor-intensive process, often requiring skilled artists to dedicate hours to crafting a single asset. This challenge is particularly acute in game and film production, where thousands of high-quality textured models are essential for constructing complex scenes and conveying compelling narratives.

Despite rapid progress in generative modeling—-driven by unprecedented advances in data availability and computational resources—-which has transformed fields such as natural language processing~\cite{achiam2023gpt}, image and video synthesis~\cite{betker2023improving,saharia2022photorealistic,wan2025wan,kong2024hunyuanvideo,blattmann2023stable}, and 3D shape generation~\cite{hong2023lrm,li2023instant3d,tochilkin2024triposr}, progress in 3D texture generation has lagged significantly. A key bottleneck is the scarcity of large-scale, high-quality 3D datasets with consistent UV parameterizations and diverse, realistic textures. Early methods~\cite{siddiqui2022texturify, bokhovkin2023mesh2tex, Yu_2023_ICCV, cheng2023tuvf} relied on curated datasets like ShapeNet~\cite{chang2015shapenet}, which suffer from limited scale and narrow category diversity, thereby restricting their generalizability. More recent efforts, such as TEXGen~\cite{yu2024texgen}, have introduced hybrid neural architectures to better model the UV space and leverage larger training sets. However, data limitations continue to hinder their generalization to diverse real-world scenarios.

Given the remarkable ability of image and video generative models to synthesize rich and diverse textures, a natural direction is to adapt these models for 3D texture generation. However, their learned priors operate in the projected 2D image space, making direct application to UV texture mapping challenging. The UV space represents an unwrapped surface manifold where pixel adjacency is governed by UV seams rather than 3D spatial continuity, introducing discontinuities. Consequently, most existing approaches generate multi-view images~\cite{huang2024mv,zeng2024paint3d,bensadoun2024meta}, followed by iterative back-projection and blending~\cite{huang2024mv,zeng2024paint3d,bensadoun2024meta}, or additional UV inpainting~\cite{zeng2024paint3d,bensadoun2024meta}, to produce complete UV texture maps. These multi-stage pipelines are not end-to-end and are prone to error accumulation and spatial inconsistencies on the textured surface.

In this paper, we present \textbf{SeqTex}, a novel end-to-end framework that harnesses the rich visual priors of pre-trained video foundation models to directly generate complete UV texture maps. Instead of predicting the UV map in isolation, we reformulate the task as a sequence generation problem: the model outputs a series of images comprising multi-view renderings of the object, with the UV texture map generated as the final frame. This design enables joint optimization of multi-view synthesis and UV texture generation within a single-stage pipeline. \textbf{SeqTex} offers three key advantages:
\textbf{(1)} By aligning the task with the temporal structure inherent in video foundation models, it effectively transfers learned visual knowledge from video data to the texture domain;
\textbf{(2)} By incorporating multi-view context, the model learns to integrate and align information across different viewpoints, resulting in more coherent and realistic UV textures;
\textbf{(3)} The unified architecture allows for training with additional high-quality multi-view-only datasets, enhancing the model’s generalization and robustness.

\textbf{SeqTex}, built upon a pre-trained video diffusion transformer, introduces three key innovations to enable high-quality 3D texture synthesis: 
{\textbf{(1)} Decoupled Multi-View (MV) and UV Texture Learning}: To bridge the domain gap between spatially coherent multi-view (MV) frames and the discontinuous layout of UV maps, we design separate processing branches. The MV branch efficiently adapts video priors using LoRA, while the UV branch is fully fine-tuned to generate high-fidelity texture maps.
{\textbf{(2)} Geometry-Informed Attention}: We leverage the geometric consistency of attributes such as 3D coordinates and surface normals across the MV and UV domains. By embedding this geometric information into the attention mechanism, we guide UV tokens to attend to MV tokens corresponding to spatially proximate 3D locations and orientations, enabling effective information transfer from the MV to the UV domain. 
{\textbf{(3)} Adaptive Token Resolution}: To capture fine-grained texture details in UV maps without incurring excessive computational overhead, we introduce an adaptive resolution strategy: UV tokens are processed at a higher spatial resolution than MV tokens. This design preserves texture details while maintaining computational efficiency.

As illustrated in Fig.~\ref{fig:teaser}, our approach effectively transfers world knowledge from pre-trained video foundation models to the UV texture space, enabling the generation of diverse a nd realistic 3D scenes with strong generalization capabilities. Our key contributions are summarized as follows: 
\begin{itemize}
\item We propose a novel end-to-end framework that adapts pre-trained video generative models to synthesize complete 3D texture maps. To the best of our knowledge, this is the first method of its kind. 
\item We introduce a decoupled architecture with dedicated branches for multi-view rendering and UV texture generation, incorporate geometry-informed attention to enhance cross-view and cross-domain information transfer, and employ adaptive resolution to efficiently capture fine-grained texture details. 
\item Our method achieves state-of-the-art performance on both image-conditioned and text-conditioned 3D texture generation tasks, consistently surpassing prior methods in 3D consistency, geometric alignment, and visual fidelity, while maintaining competitive inference speed.
\end{itemize}

\section{Related Work}
\noindent\textbf{Texture Generation Methods.} Recent years have seen rapid advances in generative modeling for 3D texture synthesis, yet key limitations persist in scalability, detail fidelity, and real-world generalization. A predominant paradigm involves leveraging pre-trained 2D diffusion or flow-matching models at test time~\cite{poole2022dreamfusion, latentnerf, fantasia3d, lin2023magic3d, wang2023prolificdreamer, yeh2024texturedreamer, richardson2023texture, chen2023text2tex, cao2023texfusion, yu2023text}, which optimize textures via image-space supervision or view-by-view inpainting. While effective in certain cases, these methods often suffer from long optimization times, lack holistic 3D awareness, and are prone to inter-view inconsistencies due to missing global context. Another line of research focuses on training feed-forward generative models directly on 3D data. Early approaches~\cite{siddiqui2022texturify, bokhovkin2023mesh2tex, Yu_2023_ICCV, cheng2023tuvf} were restricted to category-specific datasets~\cite{chang2015shapenet}, limiting their generalization to diverse, real-world assets. More recent efforts~\cite{yu2024texgen} have attempted to address this by using larger training sets and hybrid network architectures for end-to-end texture synthesis. Nevertheless, data scarcity remains a fundamental barrier; available 3D texture data is orders of magnitude smaller than the datasets used to train leading image or video models, which restricts robust generalization to open-world scenarios.

\noindent\textbf{Video Generative Models.} Video generation models~\cite{blattmann2023stable,wan2025wan,kong2024hunyuanvideo,yang2024cogvideox} have recently achieved remarkable progress, enabling the synthesis of high-quality, unprecedentedly realistic videos. Trained on large-scale video datasets, these models acquire rich world knowledge, which has been successfully adapted for downstream tasks such as controllable video generation~\cite{chen2023control,wang2023videocomposer,wan2025wan,yu2025trajectorycrafter}, video and image editing~\cite{yu2025objectmover,chen2024unireal,liu2024generative}, multi-view generation~\cite{zuo2024videomv,voleti2024sv3d}, and geometry prediction~\cite{hu2024depthcrafter,zhang2024world}. These applications benefit from operating in the projected image space, where learned priors can be directly applied. However, for 3D texture generation, the target representation lies in the UV space, which is fundamentally different from natural images, making it challenging to leverage this world knowledge. In this work, we propose a novel method that adapts pre-trained video generative models to synthesize complete 3D texture maps in an end-to-end manner. To the best of our knowledge, this is the first work to utilize video models for direct UV texture generation.
\begin{figure*}[t]
\centering
\includegraphics[width=\textwidth]{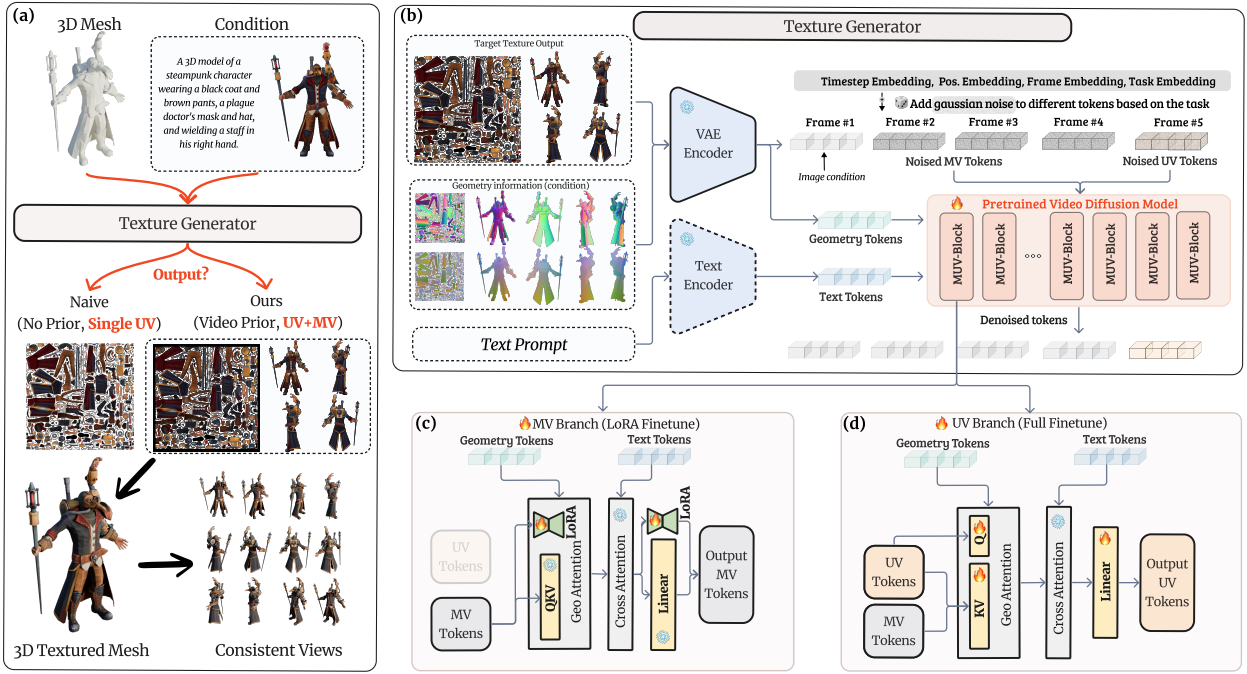} 
\caption{\textbf{Overview of our approach and core insights.} 
\textbf{(a)} Joint UV and multi-view synthesis with video priors: Given an untextured 3D mesh and conditioning inputs (image or text), SeqTex generates a complete UV texture map. Unlike prior works that predict only a single texture map, our approach jointly synthesizes multi-view images and the UV texture, thereby leveraging the rich generative priors of pre-trained video models. 
\textbf{(b)} SeqTex pipeline: Geometric image and text conditions are tokenized and injected alongside noised multi-view and UV tokens. A flow matching model—initialized from a pre-trained video foundation model and then fine-tuned for our task—denoises these tokens to yield coherent multi-view renderings and a high-fidelity UV map. 
\textbf{(c)} Multi-View (MV) branch (LoRA fine-tuned): MV tokens are refined under geometric and textual guidance. Geometry-informed attention is applied exclusively among MV tokens to enforce view consistency and produce a set of aligned multi-view outputs.
\textbf{(d)} UV branch (fully fine-tuned): Guided by geometric cues, UV tokens (queries) attend to both MV and UV tokens (keys and values) via Geo Attention. This enables the network to integrate multi-view information into a coherent UV space for accurate and seamless texture synthesis. 
}
\label{fig:pipeline}
\end{figure*}

\section{SeqTex}
\subsection{Overview}
Given an untextured 3D mesh, we aim to design a single-stage, feed-forward network-- \textbf{SeqTex} (see Figure~\ref{fig:pipeline})-- that synthesizes high-fidelity UV texture maps conditioned on a reference image and, optionally, a textual prompt, all in a single end-to-end pass. Our key insight is to harness the generative power of a pre-trained video diffusion model and adapt it for UV texture synthesis. To achieve this, we decompose the task into two subproblems-- multi-view image synthesis and UV texture mapping-- and recast their joint prediction as a sequence generation problem, thereby enabling the direct application of video diffusion models.

This formulation offers two major advantages. First, by predicting multi-view images alongside the UV texture map, SeqTex capitalizes on the strong frame synthesis capabilities of video models and uses these synthesized views to guide coherent UV texture generation. This design also enables the incorporation of large, multi-view-only datasets during training, further enhancing generalization. Second, treating the entire process as a sequential prediction task permits unified, end-to-end optimization, mitigating the error accumulation common in multi-stage pipelines.

The remainder of this section is organized as follows. Section~\ref{sec:tex} explains how we repurpose a pre-trained video diffusion model for our task. Section~\ref{sec:architecture} introduces our core architectural innovations for handling the hybrid multi-view and UV sequence representation. Finally, Section~\ref{sec:multitask} details our multi-task training strategy.

\subsection{Video Diffusion Model for UV Texture Synthesis}
\label{sec:tex}
To integrate multi-view image synthesis and UV texture mapping within a unified framework, we adapt a pre-trained transformer-based video diffusion model~\cite{wan2025wan}. We treat the sequence of multi-view and UV frames as a ``video,'' encoding each frame into the latent space using the VAE~\cite{rombach2022high} associated with the video model. Unless otherwise stated, all subsequent operations are performed in this latent domain.

At each diffusion timestep $t$, we construct a noised token sequence:
\begin{equation}
S^t = [I_1^t, I_2^t, I_3^t, I_4^t, U^t],
\end{equation}
where $I_{1:4}^t$ are the noised multi-view image latents and $U^t$ is the noised UV texture latent. Following rectified flow matching~\cite{esser2024scaling}, these latents are obtained by linearly interpolating between the clean latents $S^1 = [I_{1:4}^1, U^1]$ and Gaussian noise samples $X^0 \sim \mathcal{N}(0, \mathbf{I})$:
\begin{equation}
I_i^t = t I_i^1 + (1 - t) X_i^0 ~~(i \in \{1,2,3,4\}),\quad U^t = t U^1 + (1 - t) X_U^0.
\end{equation}
To provide 3D structural guidance and establish correspondence between frames, we condition the model on geometric cues ($c_{\text{geo}}$) derived from the input mesh, represented in both multi-view and UV domains. Unlike ISA4D~\cite{shao2025interspatial}, which employs normalized device coordinates (NDC) as an intermediary, our $c_{\text{geo}}$ comprises global positions (i.e., global coordinates) and global normals. These global attributes provide consistent geometric information across both the multi-view and UV domains, enabling finer-grained guidance. Among these two conditions, as discussed in~\cite{bensadoun2024meta}, global coordinates serve as coarse-grained indicators, conveying the overall outline of the target object, while global normals provide detailed cues to guide the generation of fine-grained results. To align with the latent tokens, these geometric features are also encoded into the latent space. Additionally, we incorporate optional textual guidance ($c_{\text{text}}$).

For training, we employ a rectified flow objective at each sampled timestep $t$:
\begin{equation}
\mathcal{L}=\mathbb{E}_{X^0, S^1, c_{\text{geo}}, c_{\text{text}}, t}\left[\|u(S^t, c_{\text{geo}}, c_{\text{text}}, t;\theta)-V^t\|^2\right],
\end{equation}
where the velocity target $V^t$ represents the direction from the noised latent to the clean latent. Its formulation is:
\begin{equation}
V^t = [I_1^1 - X_1^0, \dots, I_4^1 - X_4^0, U^1 - X_U^0].
\end{equation}

At inference, the model iteratively denoises a pure noise input to simultaneously generate coherent multi-view images and a high-fidelity UV texture map in the latent space. These latent outputs are then projected back to the pixel space using the VAE decoder.

\subsection{SeqTex Architecture}
\label{sec:architecture}
Built upon a pre-trained video diffusion transformer, we introduce the MUV Block, which is a modified transformer block tailored for UV texture synthesis. It preserves the strong multi-view generation priors of the foundation model.

\vspace{0.1in}\noindent\textbf{Decoupled MV and UV Processing Branches.} To synthesize high-quality multi-view (MV) images and a precisely aligned UV texture map in a single pass, it is crucial to bridge the domain gap between the MV space and the UV space. MV images preserve the spatial continuity of the 3D projection, whereas the UV space unwraps the surface, potentially scattering adjacent 3D points across the map. Sharing parameters, especially in normalization layers, can lead to feature interference and suboptimal learning. We therefore split the architecture into two dedicated streams: an MV branch and a UV branch, which interact via the Geo Attention through $c_{\text{geo}}$ within each MUV block.

In the MV branch (Fig.~\ref{fig:pipeline}(c)), we fine-tune only lightweight Low-Rank Adaptation (LoRA) modules while keeping the base video diffusion transformer frozen. This strategy preserves the powerful priors of the pre-trained model for multi-view synthesis and dramatically reduces computational overhead. The MV branch ingests only multi-view tokens, forming a specialized network for generating high-quality, view-consistent images. 
To inject 3D geometric guidance for the MV branch, we condition on clean position and normal maps $c_\text{geo}$ via Geo-attention:

\begin{align}\label{eq:attention}
\text{Geo-Attention}(Q,K,V,c_{\text{geo}})=\text{Softmax}(\frac{Q' K'^T}{\sqrt{d_k}})V,\\
Q'=\text{RoPE}(Q)+c_{\text{geo}}, \quad K'=\text{RoPE}(K)+c_{\text{geo}},
\end{align}
where $Q, K, \text{ and }V$ are linear-projected query, key and value tokens within the attention module:
\begin{align}
Q=\text{Linear}_q(\text{token}_q),\quad
K=\text{Linear}_k(\text{token}_e),\quad
V=\text{Linear}_v(\text{token}_e).
\end{align}
Note that both $\text{token}_q$ and $\text{token}_e$ represent the MV tokens in this branch.

In the UV branch (Fig. \ref{fig:pipeline}(d)), we fully fine-tune the backbone rather than using LoRA, since UV texture synthesis differs substantially from the original image/video-generation task that the pre-trained model was optimized for. This full fine-tuning allows the network to adapt its weights to handle the unique challenges of generating UV textures. We also integrate Geo-Attention into the UV branch, mirroring the architecture used in the MV branch. The critical distinction lies in how we build the token streams: $\text{token}_q$ contains only the UV tokens, while $\text{token}_e$ temporally concatenates both MV and UV tokens. By injecting geometric guidance—- position and normal maps—- into queries and keys, Geo-Attention steers UV tokens to focus on relevant regions in both multi-view and UV tokens. This mechanism allows UV tokens to selectively attend to geometrically aligned MV tokens, effectively narrowing the domain gap between image and UV spaces. As a result, the model can transfer the strong spatial priors learned in the MV branch into accurate, coherent UV textures, ensuring that the generated UV map aligns precisely with the synthesized multi-views.

\vspace{0.1in}\noindent\textbf{Adaptive Token Resolution.} To further reduce computation without sacrificing visual fidelity, we adopt an adaptive token resolution strategy. UV textures, where fine-grained detail is paramount, are processed at a higher spatial resolution ($H_{UV} \times W_{UV}$), while multi-view images are generated at a lower resolution ($H_{MV} \times W_{MV}$). By downsampling MV tokens, we substantially reduce computational cost, while the elevated resolution of the UV branch ensures that critical texture details are preserved.

We also adapt our positional encodings for this mixed-resolution setup. In the temporal dimension, multi-view tokens are assigned positions 1-4, and the UV token receives position 5. For the spatial dimensions, RoPE frequencies retain the native scaling strategy of the pre-trained video model, which is interpolation scaling for CogVideoX~\cite{yang2024cogvideox} and extrapolation for Wan2.1~\cite{wan2025wan}, to maximally preserve its learned priors. Further implementation details are provided in the supplementary material.

\subsection{Multi-task Learning and Material Representation}
\label{sec:multitask}
\noindent\textbf{Multi-task Learning.} Our framework supports training multiple tasks within a unified sequence representation. We define two primary tasks: image-to-texture (img2tex) and geometry-to-multi-view (geo2mv). We adopt a flexible conditioning strategy~\cite{xiao2025worldmem,chen2024diffusion} that assigns distinct noise levels to different frames based on the task. Frames are categorized as:
\begin{itemize}
    \item \textbf{Denoising Frames (DF):} Frames the model must denoise, with a noise level sampled from a predefined schedule.
    \item \textbf{Conditioning Frames (CF):} Frames provided as known information, assigned a minimal noise level.
    \item \textbf{Nonsense Frames (NF):} Frames irrelevant to the current task, assigned the maximum noise level (effectively pure noise).
\end{itemize}
This scheme ensures a unified operation for all tasks. The specific frame assignments for each task are detailed in Table~\ref{tab:frames}. The ``geo2mv'' task enables the seamless integration of additional multi-view-only datasets, as the UV frame is treated as a nonsense frame, thereby mitigating the issue of 3D data scarcity.

\begin{table}[!ht]
\centering
\caption{\textbf{Frame type assignments and noise scheduling for different tasks.}}
\label{tab:frames}
\begin{tabular}{l|lll}
\toprule
\textbf{Task} & \textbf{DF (Denoising)} & \textbf{CF (Condition)} & \textbf{NF (Nonsense)} \\
\midrule
img2tex   & MV images (excl. $I_i$), UV map ($U$) & MV image $I_i$ & None \\
geo2mv    & All MV images ($I_{1:4}$)    & None                   & UV map ($U$)  \\
\bottomrule
\end{tabular}%
\end{table}

\vspace{0.1in}\noindent\textbf{Material Representation.} 
\label{subsec:material_rep}
In recent texture generation research, two primary choices exist for material representation during training and generation: training data \textit{with} lighting (baked-in illumination) and data \textit{without} lighting (typically generating albedo, and potentially additional maps like roughness and metallicity).

Given our dual training tasks-- \textit{img2tex} and \textit{geo2mv}-- we adopt distinct material representations for each task. For \textit{img2tex}, we utilize \textbf{lighting-free maps}, specifically \textbf{albedo maps}. We posit that albedo data facilitates consistency across different viewpoints and potentially across different domains.

Conversely, for \textit{geo2mv}, we employ \textbf{illuminated images} as training data. This decision is driven by two key considerations:
\begin{itemize}
    \item \textbf{Compatibility with video foundation models:} Natural video data inherently contains lighting. Using illuminated images helps mitigate model degradation by aligning with the data distribution these foundation models are trained on.
    \item \textbf{Broader Dataset Compatibility:} Illuminated images enable the use of a wider range of multi-view datasets, including real-world scanned model datasets, and ensure compatibility with potential future datasets comprising pure video sequences.
\end{itemize}
\section{Experiments}
We use the dataset from TEXGen~\cite{yu2024texgen} for training, which contains 120,400 textured meshes: 120,000 for the training set and 400 for the evaluation set. Unlike TEXGen, we generate diverse caption styles using Claude~\cite{claude} based on multi-view renderings-- including detailed descriptions, concise summaries, and tags-- to comprehensively capture the various aspects of each model. Given the complexity and time-intensive nature of processing textured 3D data (e.g., UV parameterization and texture baking), we further curate a multi-view (MV) dataset from multiple sources, including Objaverse-XL~\cite{deitke2023objaversexl}, DTC~\cite{dong2025digital}, RenderPeople~\cite{renderpeople}, Vroid~\cite{chen2023panic3d}, and THuman~\cite{tao2021function4d}, to support training via the geo2mv task. We apply filters to ensure high aesthetic quality in both geometry and texture, yielding approximately 80,000 3D models. For these models, we render only their geometry and RGB values in the pixel domain, without UV re-parameterization or texture baking. Notably, while the 3D texture dataset contains only albedo maps, we incorporate comprehensive material information (such as metallic and roughness) using Physically Based Rendering (PBR). This ensures that the images generated by the MV branch better align with the prior distribution of the video model. To balance performance and computational efficiency, we set the resolution to $512 \times 512$ pixels for MV and $1024 \times 1024$ pixels for UV. Further training details are provided in the supplementary material.

\subsection{Image-Conditioned Texture Generation}
Following~\cite{yu2024texgen}, our primary evaluations and comparisons focus on image-conditioned generation, where both a text prompt and a pose-aligned view image are provided as inputs. We compare our method against existing texture generation approaches, including TEXGen~\cite{yu2024texgen}, TEXTure~\cite{richardson2023texture}, Text2Tex~\cite{chen2023text2tex}, and Paint3D~\cite{zeng2024paint3d}.

\begin{figure*}[htp]
\centering
\includegraphics[width=\textwidth]{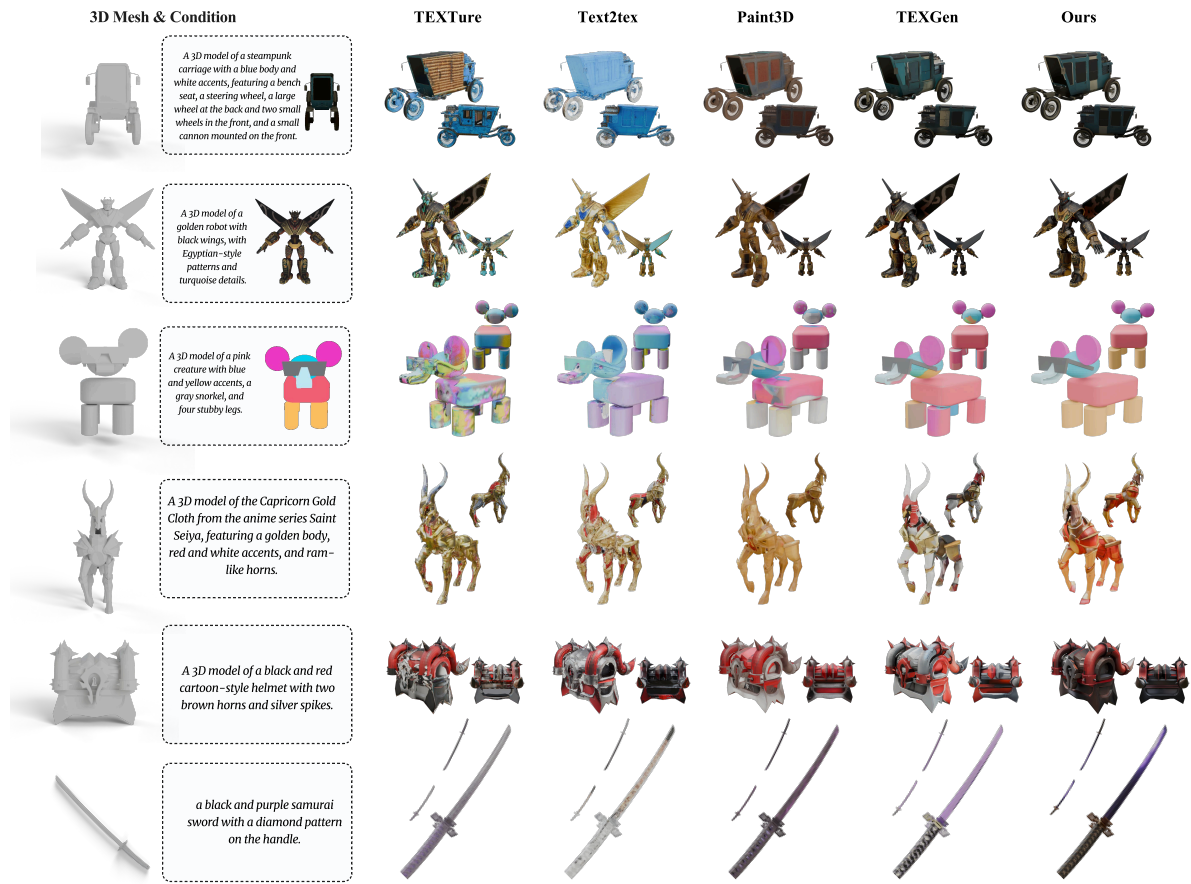} 
\caption{\textbf{Qualitative comparison of texture generation methods.} To ensure the generated textures align with the ground-truth, we adapt TEXTure, Text2Tex, and Paint3D to accept an additional image condition, following the methodology of TEXGen~\cite{yu2024texgen}.}
\label{fig:qualitative_results}
\end{figure*}

Qualitative comparisons are presented in Fig.~\ref{fig:qualitative_results}. As an end-to-end texture generation method operating directly in the UV domain, our approach produces textures that align better with the input conditions and exhibit higher fidelity than those from TEXTure and Text2Tex. Our method typically generates textures with more intricate patterns compared to the oversmoothed results from Paint3D. Relative to TEXGen, our technique demonstrates a better understanding of semantic information across different object parts and maintains improved consistency across views.

For quantitative evaluation, we assess synthesis quality on a validation set of 400 object instances. Following~\cite{yu2024texgen}, we compute two established perceptual metrics---Fréchet Inception Distance (FID) and Kernel Inception Distance (KID)---using ground-truth renderings as the reference. As shown in Table~\ref{table:quan}, our method achieves statistically significant improvements in synthesis quality over existing approaches, while exhibiting computational efficiency comparable to TEXGen under identical hardware configurations.



\begin{table}[ht]
\centering
\caption{\textbf{Quantitative results for image-conditioned generation.} FID and KID ($\times 10^{-4}$) are evaluated between multi-view renderings and ground-truth images. Our method achieves state-of-the-art texture quality with competitive inference speed.}
\vspace{-2mm}
\begin{tabular}{crrrr}
\toprule
Methods & FID($\downarrow$) & KID($\downarrow$) & Time($\downarrow$) \\
\midrule
TEXTure \cite{richardson2023texture} & 48.31 & 48.00 & 80s \\
Text2Tex \cite{chen2023text2tex} & 49.85 & 47.38 & 344s \\
Paint3D \cite{zeng2024paint3d} & 43.55 & 25.73 & 95s \\
TEXGen~\cite{yu2024texgen} & \ul{34.53} & \ul{11.94} & \textbf{10s} \\
Ours & \bf{30.27} & \bf{1.21} & \ul{12s} \\
\bottomrule
\end{tabular}
\vspace{-2mm}
\label{table:quan}
\end{table}

\subsection{Text-Conditioned Texture Generation}
To integrate text-conditioned inputs into our image-based pipeline, we first generate image conditions from user-provided text prompts using generators like SDXL~\cite{podell2023sdxl} or FLUX~\cite{flux2024}. These generators, combined with ControlNet~\cite{zhang2023adding}, process geometric cues (i.e., depth and normal maps from a specific viewpoint) to produce high-quality, detailed single-view images. These synthesized images then serve as input conditions for our texture generation pipeline, which remains unchanged. Visualizations in Fig.~\ref{fig:viz_dtc} demonstrate our method's exceptional generalization and condition-adherence capabilities.

\begin{figure*}[htp]
\centering
\vspace{-3em}
\includegraphics[width=\textwidth]{imgs/dtc.pdf} 
\caption{\textbf{Visualization of text-conditioned texture generation on the real-scan DTC dataset.} Given an untextured mesh, a text prompt, and a corresponding image condition, our method creates textures that align well with the geometry. The 3D priors from the video foundation model ensure view consistency.}
\label{fig:viz_dtc}
\end{figure*}

For the quantitative evaluation of text-conditioned generation, standard metrics like FID and KID are unsuitable, as they measure correspondence to ground-truth renderings. This is problematic when a high-quality generated texture may legitimately differ from the ground truth. Therefore, we adopt the following evaluation protocol:
\begin{itemize}
    \item We conduct a user study, following TEXGen~\cite{yu2024texgen}, to compare our results against other methods and collect user preference scores.
    \item Since no established metrics exist for this generative task, we employ Claude 3.5 Sonnet~\cite{claude}, a Multimodal Large Language Model (MLLM), to provide objective scores based on texture renderings. Implementation details are in the supplementary material.
\end{itemize}
The quantitative results are presented in Table~\ref{tab:text}.

\begin{table}[ht]
\centering
\caption{\textbf{Quantitative comparison of text-conditioned 3D texture generation.} Our method achieves the highest user preference rate and MLLM score, outperforming previous approaches by a substantial margin.}
    \begin{tabular}{lrrrrrr}
    \toprule
    \text{Methods} & \text{TEXTure} & \text{Text2Tex} & \text{Paint3D} & \text{TEXGen} & \text{Ours} \\
    \midrule
    \text{Preference (\%)} ($\uparrow$) & 2.15 & 1.43 & 7.88 & 31.50 & \textbf{57.04} \\
    \text{MLLM Score} ($\uparrow$) & 63.34 & 64.48 & 63.5 & 70.42  & \textbf{74.84} \\
    \bottomrule
    \end{tabular}
\label{tab:text}
\end{table}

\subsection{Ablation Studies}

\begin{table}
\centering
\caption{\textbf{Quantitative ablation results for different experimental settings.} To reduce the cost of full training, we use a lower resolution setting ($384 \times 384$ for MV and $768 \times 768$ for UV) for rapid experimentation. 
}
    \begin{tabular}{ccc|rr}
    \toprule
    W/ Video Prior & W/ UV Sep. & Albedo 3D Data& FID($\downarrow$) & KID($\downarrow$)\\
    \toprule
    \XSolidBrush& \Checkmark & \Checkmark & 44.62 & 31.82 \\
    \Checkmark & \XSolidBrush & \Checkmark&35.62&7.59\\
    \Checkmark&\Checkmark& \XSolidBrush&41.85 & 32.59\\
    \midrule
    \Checkmark&\Checkmark& \Checkmark & \textbf{33.12} & \textbf{6.69}\\
    \bottomrule
    \end{tabular}
\label{table:num_ablation}
\end{table}

In this section, we conduct extensive ablation studies to validate the key design choices in our framework.

\begin{figure*}[htp]
\centering
\includegraphics[width=\textwidth]{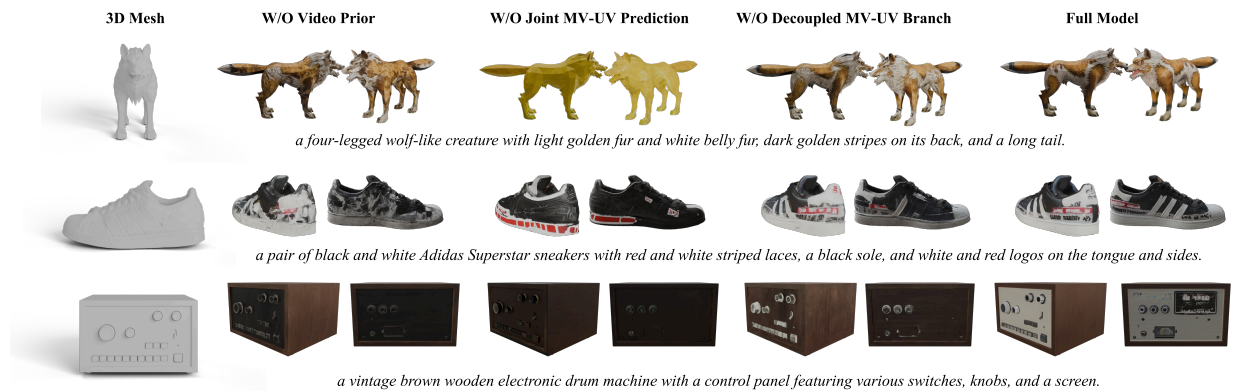} 
\caption{\textbf{Visualization of ablation study results.} Removing the video prior, joint MV-UV prediction, or the decoupled MV-UV branch leads to degraded texture fidelity, instruction-following ability, and a loss of semantic or geometric details.}
\label{fig:abl_viz}
\end{figure*}

\noindent\textbf{Effectiveness of Video Priors}
\label{sec:video_prior}
To demonstrate the importance of video priors from the pre-trained video foundation model, we train a comparative model with an identical architecture but from randomly initialized weights. As shown in Table~\ref{table:num_ablation} and Fig.~\ref{fig:abl_viz}, the randomly initialized model fails to generate high-fidelity texture maps or sufficiently detailed MV images. We also observe that fine-tuning from pre-trained weights leads to faster convergence and superior optimization, whereas the randomly initialized model tends to converge to suboptimal local minima.

\noindent\textbf{Effectiveness of Joint MV-UV Modeling}
\label{sec:joint}
The unified modeling of MV and UV frames allows the UV frame to leverage video priors from the foundation model, resulting in high-fidelity texture generation.
To validate this design, we train an alternative model that generates only UV texture maps using the UV branch, initialized from the same pre-trained weights. As evidenced by Fig.~\ref{fig:abl_viz}, the absence of MV guidance causes a significant degradation in performance, particularly in capturing details and adhering to the prompt.

\noindent\textbf{Effectiveness of Decoupled MV-UV Branch}
\label{sec:ablation_mv_uv_seperation}
Decoupling the MV and UV branches prevents interference between them, as they operate in fundamentally distinct representation domains. To assess this design choice, we trained a model without decoupled branches, where MV and UV tokens are concatenated along the temporal dimension and processed through a unified branch. The performance differences, quantified in Table~\ref{table:num_ablation}, reveal potential interference between the MV and UV branches. Furthermore, Fig.~\ref{fig:abl_viz} clearly shows that without decoupling, the model tends to lose details.

\begin{figure}[htp]
\centering
\begin{minipage}[t]{0.48\textwidth}
    \centering
    \includegraphics[width=\linewidth]{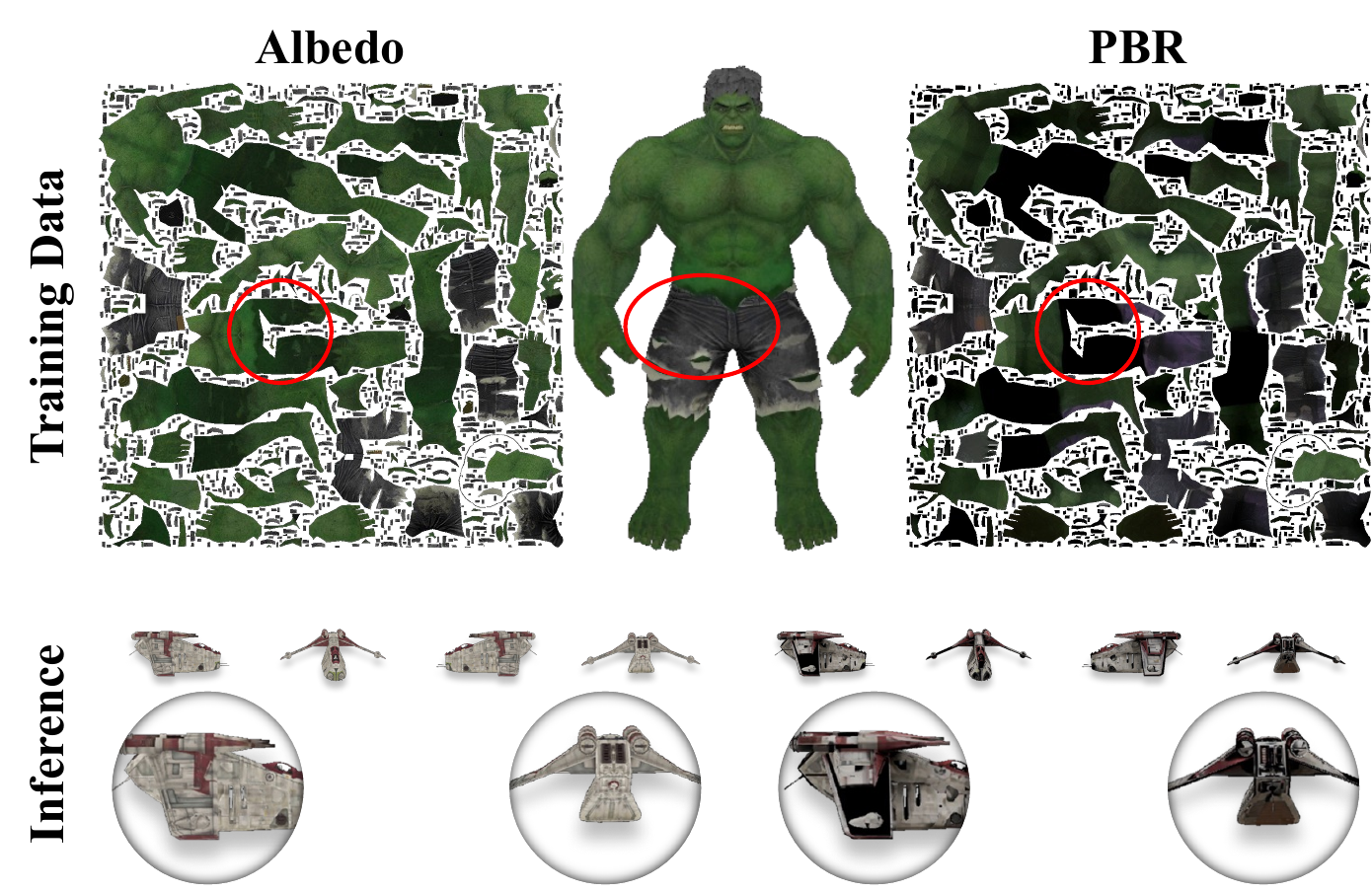}
    \caption{\textbf{Comparison of training with albedo vs. PBR texture maps.} 
    PBR texture maps contain baked-in shadows in occluded areas. 
    A model trained on such data learns to simulate these shadows, leading to incorrect artifacts during inference.}
    \label{fig:pbr_vs_albedo}
\end{minipage}
\hfill
\begin{minipage}[t]{0.48\textwidth}
    \vspace{-12em} 
    \centering
    \includegraphics[width=\linewidth]{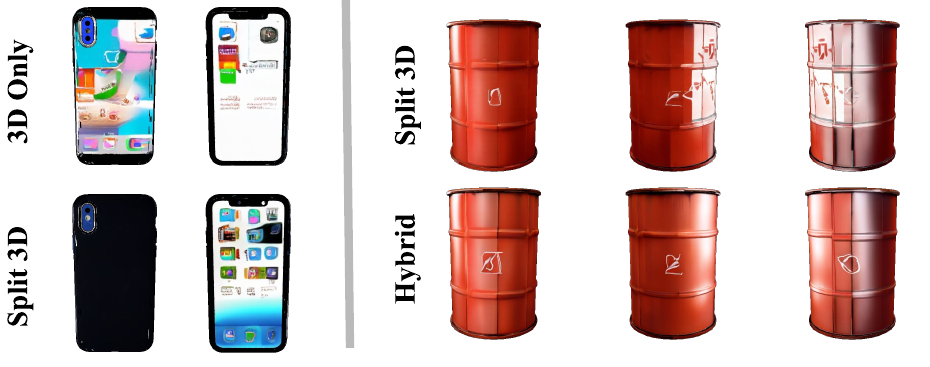}
    \caption{\textbf{Qualitative comparison of three training strategies: 
    3D-only, Split 3D, and Hybrid.} 
    The 3D-only strategy struggles with view consistency. Both Split 3D and Hybrid 
    strategies improve multi-view coherence. Hybrid training, which incorporates additional 
    PBR multi-view data, further enhances the model's understanding of lighting 
    and shadow effects.}
    \label{fig:multi_task}
    \vspace{-0.2in} 
\end{minipage}
\end{figure}

\noindent\textbf{Choice of Material Representation}
\label{sec:ablation_pbr}
We ablate the choice of material representation for the \textit{img2tex} task, with quantitative results summarized in Table~\ref{table:num_ablation}. Training \textit{img2tex} with illuminated (PBR) texture maps proves to be suboptimal. During inference, this configuration generates artifacts characterized by unnaturally dark regions. As anticipated, this occurs because illuminated (baked) textures inherently contain black regions in unlit areas (see Fig.~\ref{fig:pbr_vs_albedo}). Consequently, the model learns to replicate these shadow artifacts, producing erroneous dark patches in novel contexts where such lighting conditions do not exist.

\noindent\textbf{Effectiveness of Multi-task Training \& Dataset Composition}
We first clarify two related but distinct concepts: the geo2mv \textit{task} and the externally curated MV \textit{dataset}. 
Integrating our additional MV dataset via the geo2mv task enhances texture generation generalization. However, the geo2mv task itself does not strictly require this external dataset, as the multi-view renderings from our core 120k Objaverse models can also support this training. To systematically evaluate the efficacy of each component, we conduct controlled ablations under the following conditions:
\begin{itemize}
    \item \textbf{3D-only}: The model is fine-tuned exclusively on 3D data for the img2tex task.
    \item \textbf{Split 3D}: The 3D data is partitioned in a 3:2 ratio (approximating the 120k:80k distribution of the main training) for concurrent img2tex and geo2mv training.
    \item \textbf{Hybrid}: The model is trained on the full 3D dataset and the external MV dataset for their respective tasks.
\end{itemize}
All configurations started from a base model pre-trained for 10k iterations on the img2tex task to ensure basic convergence while conserving computational resources.

Figure~\ref{fig:multi_task} demonstrates two key findings: 1) Without the emphasis provided by the MV task (3D-only vs. Split 3D), the model struggles to maintain multi-view consistency. 2) Incorporating additional PBR MV data (Hybrid vs. Split 3D) improves the model's understanding of lighting and shadow effects, which are absent in the albedo-only data.

\section{Conclusion}
In this work, we present SeqTex, a native texture model that unlocks pretrained video foundation models for end-to-end UV texture map generation. SeqTex departs from traditional pipelines by reframing 3D texture synthesis as a joint sequence modeling task over multi-view images and UV texture maps. This approach enables the direct transfer of rich, consistent image-space priors into the UV domain, addressing long-standing challenges of data scarcity and UV spatial discontinuity.

Our architecture features a decoupled MV-UV branch design for specialized representation learning, geometry-informed attention to ensure precise alignment between the image and UV domains, and adaptive token resolution to efficiently capture fine-grained texture details while maintaining computational efficiency. These innovations allow SeqTex to effectively utilize pretrained video priors without compromising model capacity or training stability.

Extensive experiments demonstrate that SeqTex achieves state-of-the-art results on both image- and text-conditioned 3D texture generation tasks. The generated textures exhibit superior 3D consistency, accurate texture-geometry alignment, and high visual fidelity, while generalizing effectively to real-world scenarios. SeqTex establishes a strong baseline for integrating vision foundation models into practical 3D content creation pipelines and opens new avenues for scalable and robust texture synthesis.
\bibliographystyle{unsrt}  
\bibliography{bibs}

\newpage
\section{Supplementary Material}

\subsection{Additional Visualizations}
Additional visualization results are presented in Fig.~\ref{fig:text_cond_scene}.

\begin{figure*}[htp]
\centering
\vspace{-0.2em}
\includegraphics[height=0.95\textheight]{imgs/Fig1_supp.pdf} 
\caption{\textbf{Top}: An indoor scene where all objects are textured using SeqTex. \textbf{Bottom}: Close-up renderings of selected objects from the scene above.}
\label{fig:text_cond_scene}
\end{figure*}

\subsection{Implementation Details}
During training, we do not employ logit-normal time sampling~\cite{esser2024scaling}; instead, we implement a loss reweighting technique that assigns higher weights to intermediate noise timesteps and lower weights (potentially zero) to the initial and final timesteps.
We use the AdamW optimizer with a learning rate of $1 \times 10^{-4}$, betas of $0.9$ and $0.999$, and a weight decay of $0.01$. The learning rate schedule includes a 200-step warmup phase followed by cosine annealing. Our training setup consists of four nodes, each equipped with eight A800 GPUs. We use a per-GPU batch size of 1 and gradient accumulation over 4 steps, resulting in an effective global batch size of 128. We note that classifier-free guidance (CFG) is ineffective for the img2tex task, which we attribute to the stronger conditioning provided by the image input compared to the text prompt. Exponential Moving Average (EMA) is applied following the methodology in EDM2~\cite{karras2023analyzing}, with a standard deviation of $0.05$. Training is performed using bfloat16 (bf16) precision on the 1.3 billion parameter version of the Wan2.1 model. To balance performance and computational cost, we set the resolution to $512 \times 512$ for multi-view (MV) and $1024 \times 1024$ for UV. Based on insights from the Wan paper, we estimate the flow shift to be $5.0$. For inference, we use the UniPC scheduler~\cite{zhao2023unipc} with 30 steps to accelerate the process. Following~\cite{xiao2025worldmem}, during both training and inference, $k_{max}$ is set to 1000 to mask UV information for the geo2mv task, while $k_{min}$ is set to 15 (out of 1000) to provide the image condition.

The main experiment is conducted in two stages. The first stage focuses on UV generation, training the model solely on the img2tex task. The second stage introduces the geo2mv task and incorporates the new MV dataset.

\subsection{Evaluation of Text-Conditioned Texture Generation}
\begin{figure*}[h]
\centering
\includegraphics[width=\textwidth]{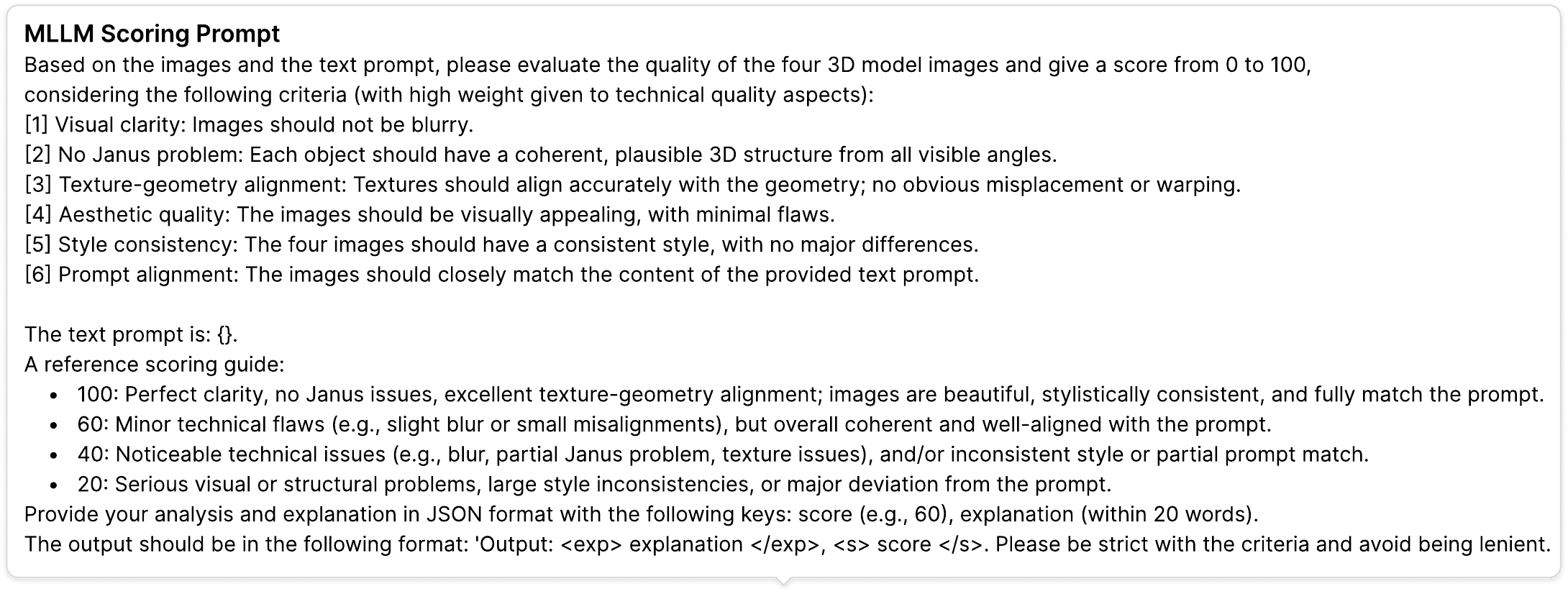}
\caption{The prompt used to evaluate text-conditioned texture generation performance.}
\label{fig:mllm_prompt}
\end{figure*}

As mentioned in the main paper, we use an MLLM score as a metric to evaluate the performance of different methods. Specifically, we use Claude 3.5 Sonnet, with the prompt shown in Fig.~\ref{fig:mllm_prompt}.

\subsection{Modified 3D RoPE}
To support different resolutions for the MV and UV inputs, as mentioned in the main text, we modify the standard 3D RoPE. The pseudo-code is provided in Alg.~\ref{alg:3d_rope_simplified}. In summary, we maintain the respective spatial resolutions of the inputs but retrieve the RoPE features for the UV map immediately following the time indices of the MV features.

\begin{algorithm}[!ht]
\caption{3D Rotary Position Embedding (RoPE) Generation}
\label{alg:3d_rope_simplified}
\SetAlgoLined
\KwIn{\texttt{input\_freqs}, \texttt{mv\_shape}, \texttt{uv\_shape}, \texttt{embed\_dim}}
\KwOut{\textbf{3D RoPE embeddings}}
\BlankLine
$(\texttt{mv\_frames}, \texttt{mv\_height}, \texttt{mv\_width}) \gets \texttt{mv\_shape}$\;
$(\texttt{uv\_frames}, \texttt{uv\_height}, \texttt{uv\_width}) \gets \texttt{uv\_shape}$\;
\BlankLine
$\texttt{d\_part} \gets \texttt{embed\_dim} / 6$\;
$(\texttt{freq\_t}, \texttt{freq\_h}, \texttt{freq\_w}) \gets \textbf{split}(\texttt{input\_freqs}, [\texttt{d\_part}, \texttt{d\_part}, \texttt{d\_part}])$\;
\BlankLine
\tcp{\textbf{Generate Multi-View RoPE components}}
$\texttt{mv\_ft} \gets \textbf{reshape}(\texttt{freq\_t}[:\texttt{mv\_frames}], (\texttt{mv\_frames},1,1,-1))$\;
$\texttt{mv\_ft} \gets \textbf{expand}(\texttt{mv\_ft}, (\texttt{mv\_frames}, \texttt{mv\_height}, \texttt{mv\_width}, -1))$\;
$\texttt{mv\_fh} \gets \textbf{reshape}(\texttt{freq\_h}[:\texttt{mv\_height}], (1,\texttt{mv\_height},1,-1))$\;
$\texttt{mv\_fh} \gets \textbf{expand}(\texttt{mv\_fh}, (\texttt{mv\_frames}, \texttt{mv\_height}, \texttt{mv\_width}, -1))$\;
$\texttt{mv\_fw} \gets \textbf{reshape}(\texttt{freq\_w}[:\texttt{mv\_width}], (1,1,\texttt{mv\_width},-1))$\;
$\texttt{mv\_fw} \gets \textbf{expand}(\texttt{mv\_fw}, (\texttt{mv\_frames}, \texttt{mv\_height}, \texttt{mv\_width}, -1))$\;
\BlankLine
\tcp{\textbf{Generate UV RoPE components}}
$\texttt{uv\_ft} \gets \textbf{reshape}(\texttt{freq\_t}[\texttt{mv\_frames}:\texttt{mv\_frames}+\texttt{uv\_frames}], (\texttt{uv\_frames},1,1,-1))$\;
$\texttt{uv\_ft} \gets \textbf{expand}(\texttt{uv\_ft}, (\texttt{uv\_frames}, \texttt{uv\_height}, \texttt{uv\_width}, -1))$\;
$\texttt{uv\_fh} \gets \textbf{reshape}(\texttt{freq\_h}[:\texttt{uv\_height}], (1,\texttt{uv\_height},1,-1))$\;
$\texttt{uv\_fh} \gets \textbf{expand}(\texttt{uv\_fh}, (\texttt{uv\_frames}, \texttt{uv\_height}, \texttt{uv\_width}, -1))$\;
$\texttt{uv\_fw} \gets \textbf{reshape}(\texttt{freq\_w}[:\texttt{uv\_width}], (1,1,\texttt{uv\_width},-1))$\;
$\texttt{uv\_fw} \gets \textbf{expand}(\texttt{uv\_fw}, (\texttt{uv\_frames}, \texttt{uv\_height}, \texttt{uv\_width}, -1))$\;
\BlankLine
\tcp{\textbf{Combine and reshape}}
$\texttt{rope\_mv} \gets \textbf{concat}([\texttt{mv\_ft}, \texttt{mv\_fh}, \texttt{mv\_fw}], \textbf{dim}=-1)$\;
$\texttt{rope\_mv} \gets \textbf{reshape}(\texttt{rope\_mv}, (\texttt{mv\_frames} \times \texttt{mv\_height} \times \texttt{mv\_width}, -1))$\;
$\texttt{rope\_uv} \gets \textbf{concat}([\texttt{uv\_ft}, \texttt{uv\_fh}, \texttt{uv\_fw}], \textbf{dim}=-1)$\;
$\texttt{rope\_uv} \gets \textbf{reshape}(\texttt{rope\_uv}, (\texttt{uv\_frames} \times \texttt{uv\_height} \times \texttt{uv\_width}, -1))$\;
\BlankLine
\Return{$\textbf{concat}([\texttt{rope\_mv}, \texttt{rope\_uv}], \textbf{dim}=0)$}\;
\end{algorithm}

\subsection{Limitations}

Although we adopt a VAE to encode geometry information for alignment with the RGB latent space, this compression inevitably introduces distortion. We find that the reconstructed geometry maps appear visually identical to the originals; however, quantitative analysis reveals non-negligible differences, which may impair the controllability of the geo attention mechanism, finally causing blurred details. 

These discrepancies highlight a limitation in our current approach and indirectly justify the necessity of high-resolution texture synthesis for preserving fine-grained geometric features. We hypothesize that the root cause lies in the domain gap between RGB images and geometry maps in the VAE's training data. Due to time constraints, we leave a more principled solution to this issue as future work.

\end{document}